\documentclass{article}

\usepackage{arxiv}

\usepackage[utf8]{inputenc}
\usepackage[T1]{fontenc}
\usepackage{hyperref}
\usepackage{url}
\usepackage{booktabs}
\usepackage{amsfonts}
\usepackage{amsmath}
\usepackage{amssymb}
\usepackage{nicefrac}
\usepackage{microtype}
\usepackage{graphicx}
\usepackage{multirow}
\usepackage{natbib}
\usepackage{doi}

\graphicspath{ {./} }

\title{A Modular Zero-Shot Pipeline for Accident Detection, Localization, and Classification in Traffic Surveillance Video}

\author{
	\href{https://orcid.org/0000-0001-5644-1575}{\includegraphics[scale=0.06]{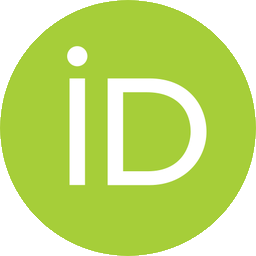}\hspace{1mm}Amey Thakur} \\
	Independent Researcher \\
	Toronto, Canada \\
	\texttt{ameythakur20@gmail.com} \\
	\And
	\href{https://orcid.org/0009-0002-0818-461X}{\includegraphics[scale=0.06]{orcid.png}\hspace{1mm}Sarvesh Talele} \\
	Independent Researcher \\
	Mumbai, India \\
	\texttt{talelesarvesh@gmail.com} \\
}

\renewcommand{\undertitle}{}
\renewcommand{\headeright}{}
\renewcommand{\shorttitle}{Zero-Shot Accident Detection Pipeline}

\hypersetup{
	pdftitle={A Modular Zero-Shot Pipeline for Accident Detection, Localization, and Classification in Traffic Surveillance Video},
	pdfsubject={cs.CV},
	pdfauthor={Amey Thakur, Sarvesh Talele},
	pdfkeywords={Zero-Shot Learning, Accident Detection, Traffic Surveillance, CLIP, Optical Flow},
}

\begin{document}
\maketitle

\begin{abstract}
We describe a zero-shot pipeline developed for the ACCIDENT @ CVPR 2026 challenge.
The challenge requires predicting when, where, and what type of traffic accident occurs in surveillance video, without labeled real-world training data.
Our method separates the problem into three independent modules.
The first module localizes the collision in time by running peak detection on z-score normalized frame-difference signals.
The second module finds the impact location by computing the weighted centroid of cumulative dense optical flow magnitude maps using the Farneback algorithm.
The third module classifies collision type by measuring cosine similarity between CLIP image embeddings of frames near the detected peak and text embeddings built from multi-prompt natural language descriptions of each collision category.
No domain-specific fine-tuning is involved; the pipeline processes each video using only pre-trained model weights.
Our implementation is publicly available as a Kaggle notebook~\citep{thakur2026notebook}.
\end{abstract}


\section{Introduction}
\label{sec:intro}

Road traffic crashes kill over one million people each year according to the World Health Organization~\citep{who2023road}. Surveillance cameras installed at intersections and along highways record continuous footage that captures the circumstances of many of these incidents. If this footage could be analyzed automatically, emergency dispatchers would receive faster alerts and investigators would gain access to objective scene reconstructions. Yet most current detection methods require supervised training on annotated video from the same deployment environment~\citep{sultani2018real, yao2022dota}, which makes them expensive to transfer across camera installations with different viewpoints, lighting, and traffic patterns.

\begin{figure}[ht]
    \centering
    \includegraphics[width=\linewidth]{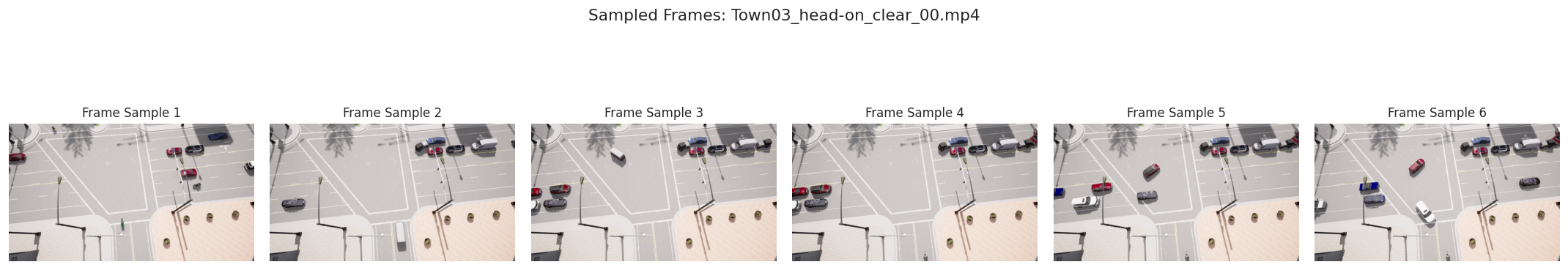}
    \caption{Chronological sampled frames from a synthetic CARLA traffic incident in the ACCIDENT @ CVPR 2026 dataset. Most clips record approximately 18 seconds of motion at 20 FPS resolution.}
    \label{fig:sampled_frames}
\end{figure}

The ACCIDENT @ CVPR 2026 competition~\citep{accident2026} tests whether accident analysis can work without any labeled real-world data. Participants receive only synthetic videos generated by the CARLA simulator~\citep{dosovitskiy2017carla} for development (see Figure~\ref{fig:sampled_frames}); the test set is composed of real CCTV recordings and manual annotation of it is prohibited.

Three predictions are required per video: (1) the accident time in seconds, (2) normalized spatial coordinates of the impact point, and (3) the collision type from a five-class taxonomy (head-on, rear-end, sideswipe, single-vehicle, t-bone). Scoring uses the harmonic mean of a Gaussian temporal similarity, a Gaussian spatial similarity, and top-1 classification accuracy, so a poor prediction in any one dimension pulls the overall score down.

We address each target with a separate module. Temporal localization runs statistical anomaly detection on pixel-intensity differences between consecutive frames. Spatial localization accumulates dense optical flow magnitudes and extracts a weighted centroid. Collision classification passes frames near the detected peak through CLIP~\citep{radford2021learning}, a contrastive vision-language model trained on 400 million image-text pairs, and selects the class whose text prompts produce the highest cosine similarity. Because the modules are independent, any one of them can be swapped or tuned without modifying the rest. The full pipeline is implemented in a public Kaggle notebook~\citep{thakur2026notebook}.

\section{Method}
\label{sec:method}

\subsection{Problem Formulation}

A test video $V$ has $N$ frames recorded at $f$ frames per second. The task is to produce a tuple $(t^{*}, c_x^{*}, c_y^{*}, k^{*})$: $t^{*}$ is the accident time in seconds, $(c_x^{*}, c_y^{*}) \in [0, 1]^{2}$ are normalized image coordinates of the point of impact, and $k^{*} \in \mathcal{K}$ is the collision type. The label space is $\mathcal{K} = \{\texttt{head-on}, \texttt{rear-end}, \texttt{sideswipe}, \texttt{single}, \texttt{t-bone}\}$.

We split the problem into three parts. Time and location predictions rely on classical signal processing; type classification relies on a pre-trained vision-language model.

\subsection{Temporal Peak Detection}

Collisions produce sudden intensity changes between consecutive frames. We exploit this observation by building a one-dimensional signal from per-frame brightness differences and then looking for statistical outliers in that signal.

Let $I_t \in \mathbb{R}^{H \times W}$ be the grayscale frame at index $t$, resized to $H = 180$, $W = 320$ for speed. We compute the mean absolute difference between adjacent frames:
\begin{equation}
    d_t = \frac{1}{HW} \sum_{u=1}^{H} \sum_{v=1}^{W} \big| I_{t+1}(u,v) - I_t(u,v) \big|
    \label{eq:framediff}
\end{equation}
for $t = 1, \ldots, N{-}1$. This series contains both collision signals and background noise from camera shake and ordinary traffic. A centered rolling mean with window $w = 5$ suppresses the short-lived noise:
\begin{equation}
    \bar{d}_t = \frac{1}{|W_t|} \sum_{s \in W_t} d_s
    \label{eq:smooth}
\end{equation}
where $W_t = \{s : |s - t| \leq \lfloor w/2 \rfloor\}$. We normalize the smoothed values into z-scores using the series-wide mean $\mu$ and standard deviation $\sigma$:
\begin{equation}
    z_t = \frac{\bar{d}_t - \mu}{\sigma + \varepsilon}
    \label{eq:zscore}
\end{equation}
with $\varepsilon = 10^{-8}$ to avoid division by zero. Any frame whose $z_t$ exceeds a threshold $\tau$ is treated as an anomaly candidate. Among all candidates, we select the one with the strongest anomaly score:
\begin{equation}
    t^{*}_{\text{frame}} =
    \begin{cases}
        \arg\max_{t :\, z_t > \tau}\, z_t & \exists\, t \text{ s.t. } z_t > \tau \\
        \arg\max_t\, z_t & \text{otherwise}
    \end{cases}
    \label{eq:peak}
\end{equation}
with $\tau = 1.5$. Taking the highest-scoring candidate among threshold-crossers favors the most prominent motion event. When no frame crosses the threshold, we fall back to the global maximum. Time in seconds is $t^{*} = t^{*}_{\text{frame}} / f$.

\begin{figure}[t]
    \centering
    \includegraphics[width=\linewidth]{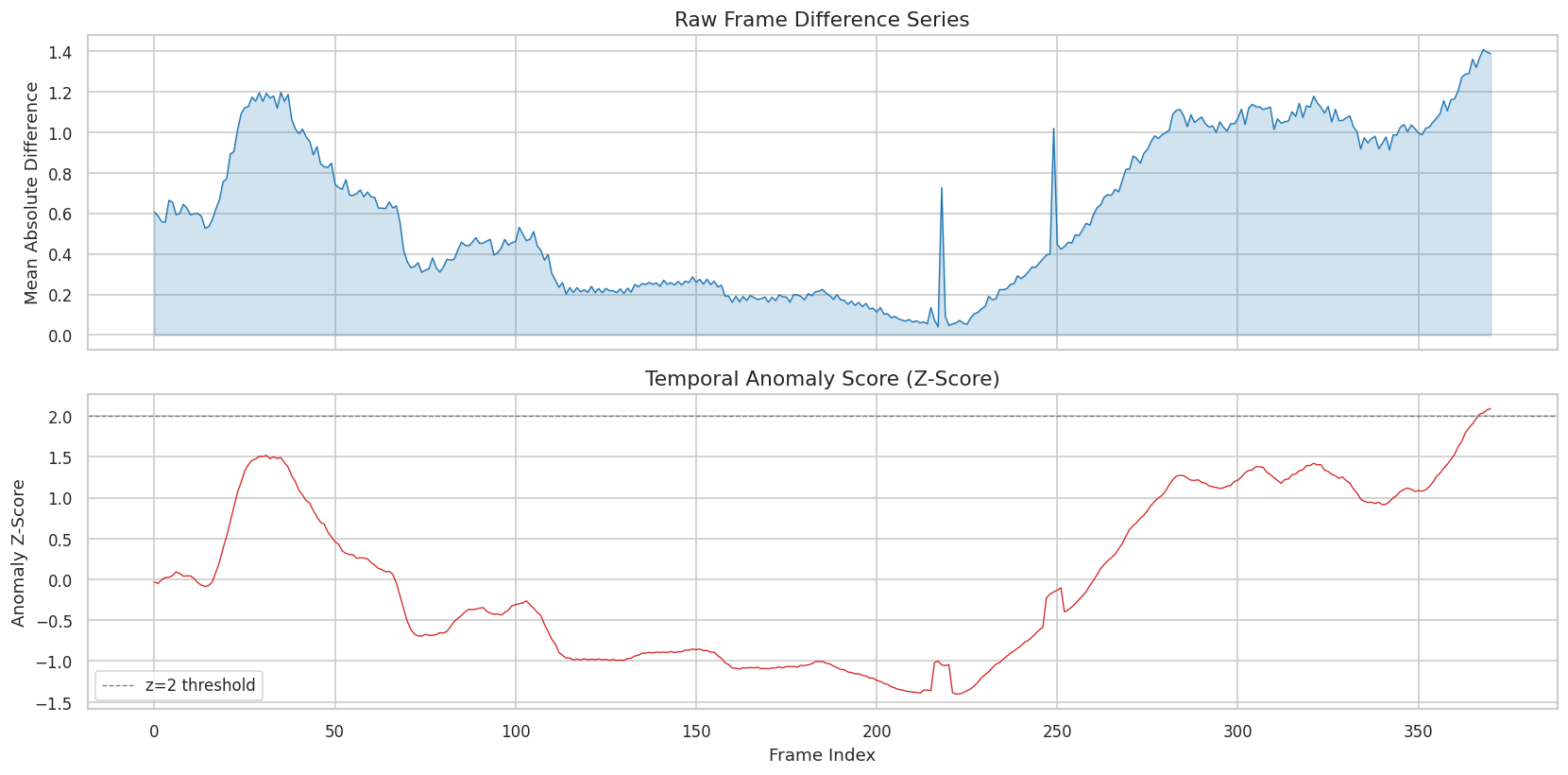}
    \caption{Temporal localization on a synthetic CARLA video. Top: mean absolute frame difference $d_t$ across all frames. Bottom: z-score anomaly series $z_t$ after rolling-mean smoothing. The dashed line marks the detection threshold.}
    \label{fig:temporal_signal}
\end{figure}

Figure~\ref{fig:temporal_signal} shows the raw frame-difference series and its z-score transform for a synthetic head-on collision. The raw signal (top panel) contains a mixture of gradual intensity drift from vehicle motion and a sharp transient near the collision moment. After smoothing and normalization (bottom panel), the collision event stands out as a distinct peak while background variation stays below the detection threshold.

\subsection{Spatial Impact Localization}

A collision concentrates high-magnitude motion in a small region of the image. We identify that region by accumulating dense optical flow over a short window and then computing the weighted centroid of the resulting magnitude map.

When the predicted accident time $t^{*}$ is available from the temporal module, we center a 30-frame window on the corresponding frame index; otherwise, we fall back to starting at frame $\lfloor N/3 \rfloor$. For each pair of consecutive frames within this window, we run the Farneback dense optical flow algorithm~\citep{farneback2003two} at $320 \times 180$ resolution. Farneback estimates per-pixel displacement using quadratic polynomial expansions and a multi-scale Gaussian pyramid. We set pyramid scale to $0.5$, three levels, window size $15$, three iterations, and polynomial parameters $n = 5$, $\sigma_p = 1.2$.

Let $\mathbf{f}_t(u,v) = (f_x, f_y)$ be the flow vector at pixel $(u,v)$ between frames $t$ and $t{+}1$. We sum the displacement magnitudes:
\begin{equation}
    M(u,v) = \sum_{t} \sqrt{f_{x}^{\,2}(u,v,t) + f_{y}^{\,2}(u,v,t)}
    \label{eq:magmap}
\end{equation}
Before extracting the centroid, we apply a percentile threshold at the 90th percentile of $M$ and zero out all values below it. This suppresses diffuse background motion and retains only the high-magnitude cluster near the collision site. The impact location is the weighted centroid of the thresholded map, normalized to unit coordinates:
\begin{equation}
    c_x^{*} = \frac{1}{W}\,\frac{\sum_{u,v} v \cdot M(u,v)}{\sum_{u,v} M(u,v)}, \quad
    c_y^{*} = \frac{1}{H}\,\frac{\sum_{u,v} u \cdot M(u,v)}{\sum_{u,v} M(u,v)}
    \label{eq:centroid}
\end{equation}
If $\sum M < 10^{-6}$, we return the frame center $(0.5, 0.5)$. This centroid calculation is a special case of the image moment framework introduced by \citet{hu1962visual}, applied to flow magnitudes instead of pixel intensities.

\begin{figure}[t]
    \centering
    \includegraphics[width=\linewidth]{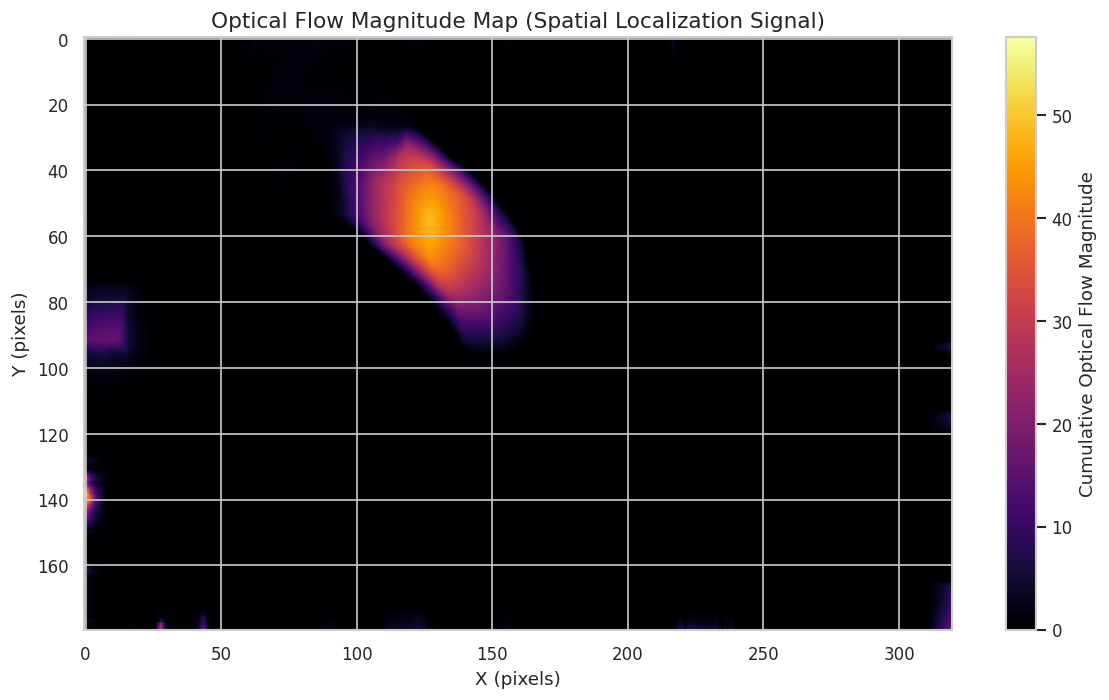}
    \caption{Cumulative Farneback optical flow magnitude $M(u,v)$ on a synthetic CARLA video after 90th-percentile thresholding. The bright region corresponds to the collision area; diffuse background motion is suppressed.}
    \label{fig:heatmap}
\end{figure}

Figure~\ref{fig:heatmap} shows the accumulation result on a synthetic video. The collision region concentrates most of the optical flow energy into a compact spatial cluster. After percentile thresholding, the weighted centroid falls within that cluster and provides a localized impact coordinate.

\subsection{Collision Type Classification}

We assign a collision type using CLIP~\citep{radford2021learning}, which learns a joint embedding space for images and text by contrastive training on 400 million image-text pairs collected from the internet. At test time, CLIP compares an image embedding against text embeddings of candidate class names using cosine similarity, allowing classification without any task-specific training data.

For each type $k \in \mathcal{K}$, we write five natural language descriptions $\{p_k^1, \ldots, p_k^5\}$ that characterize the collision from a bystander perspective (Table~\ref{tab:prompts}). Using several prompts per class reduces the influence of any one particular wording, consistent with the prompt ensembling approach studied by \citet{radford2021learning} and \citet{zhou2022learning}. Each prompt is encoded by the CLIP text encoder $\phi_{\text{text}}$, L2-normalized, and then averaged:
\begin{equation}
    \mathbf{t}_k = \frac{1}{5} \sum_{j=1}^{5} \frac{\phi_{\text{text}}(p_k^j)}{\|\phi_{\text{text}}(p_k^j)\|}
    \label{eq:textfeat}
\end{equation}
These vectors are computed once and cached before processing any video.

At inference time, we extract eight video frames centered on the predicted accident time $t^{*}$ and feed them into the CLIP visual encoder (ViT-B/32). Each embedding is L2-normalized, and we average the eight into a single representation:
\begin{equation}
    \mathbf{v} = \frac{1}{8} \sum_{i=1}^{8} \frac{\phi_{\text{img}}(I_{t_i})}{\|\phi_{\text{img}}(I_{t_i})\|}
    \label{eq:imgfeat}
\end{equation}
The predicted type is the class with highest similarity:
\begin{equation}
    k^{*} = \arg\max_{k \in \mathcal{K}} \; \mathbf{v} \cdot \mathbf{t}_k
    \label{eq:classify}
\end{equation}

\begin{table}[t]
\centering
\small
\caption{Prompt templates for collision type classification. Five prompts per class are averaged into one text embedding.}
\label{tab:prompts}
\begin{tabular}{@{}ll@{}}
\toprule
\textbf{Type} & \textbf{Example prompt} \\
\midrule
head-on   & ``two cars colliding head-on from opposite directions'' \\
rear-end  & ``a car colliding into the back of another car'' \\
sideswipe & ``two vehicles scraping alongside each other'' \\
single    & ``a single car crashing into a wall or obstacle'' \\
t-bone    & ``a car hitting the side of another car at an intersection'' \\
\bottomrule
\end{tabular}
\end{table}

\section{Experiments}
\label{sec:experiments}

\subsection{Dataset}

\begin{figure}[t]
    \centering
    \includegraphics[width=\linewidth]{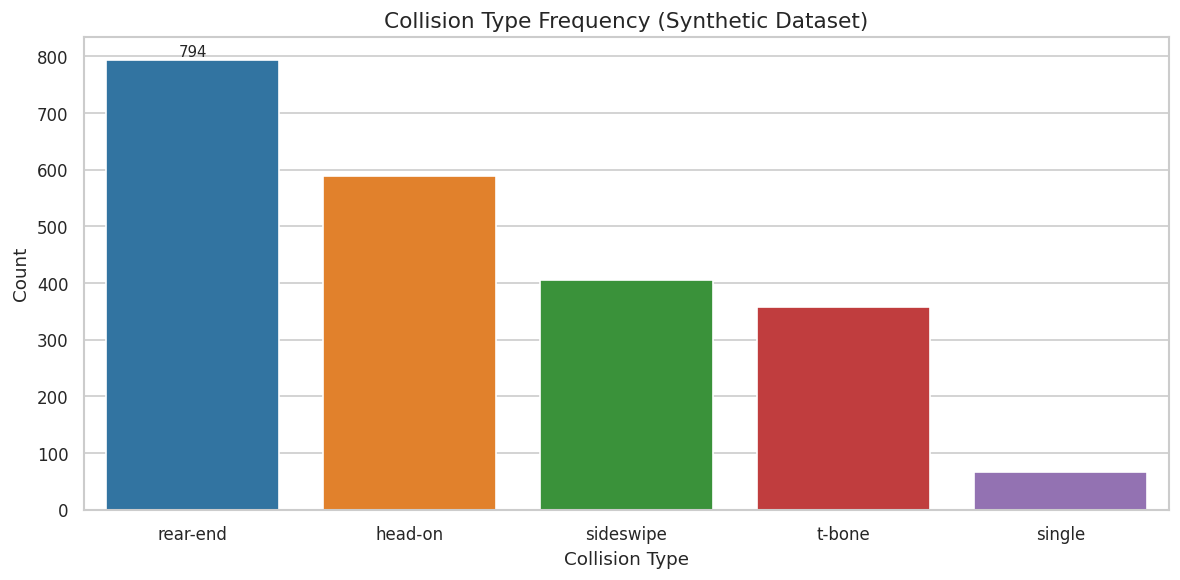}
    \caption{Collision type frequency in the synthetic development split. The rear-end category has 12 times more samples than the single-vehicle category.}
    \label{fig:type_freq}
\end{figure}

The ACCIDENT @ CVPR 2026 dataset~\citep{accident2026} has two splits. The development split contains 2,211 synthetic CCTV-style videos rendered with the CARLA simulator~\citep{dosovitskiy2017carla} across five collision categories, each annotated with accident time, impact coordinates, and collision type. Figure~\ref{fig:type_freq} shows the class distribution: rear-end collisions are the most frequent category (794 videos), followed by head-on (588), sideswipe (405), t-bone (358), and single-vehicle (66). The test split contains 2,027 real surveillance recordings collected from public traffic camera feeds. Test videos vary in resolution, frame rate, and lighting. Compression artifacts, lens distortion, and partial occlusion are common. Participants may not annotate the test split by hand, so any method that runs on it must generalize from the synthetic domain or from general-purpose pre-training alone.

\begin{figure}[t]
    \centering
    \includegraphics[width=\linewidth]{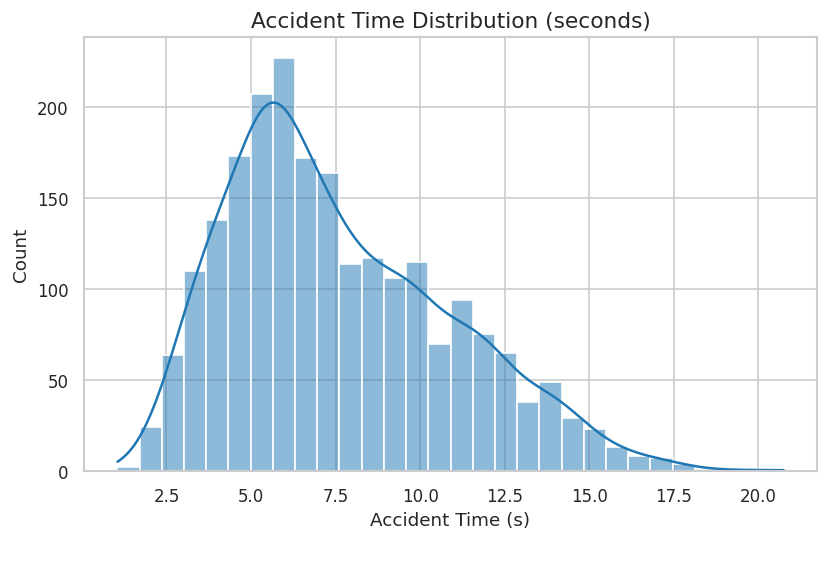}
    \caption{Distribution of ground-truth accident times (seconds) across the 2,211 synthetic videos. Most incidents occur within the first 10 seconds of the clip.}
    \label{fig:time_dist}
\end{figure}

\begin{figure}[t]
    \centering
    \includegraphics[width=\linewidth]{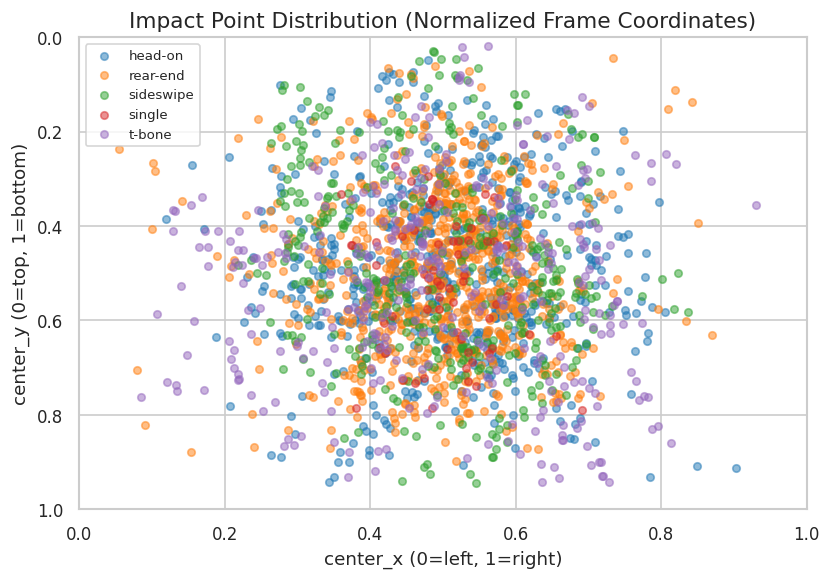}
    \caption{Ground-truth impact point distribution across 2,211 synthetic videos, colored by collision type. Points cluster near the frame center, with head-on and sideswipe events showing the widest spatial spread.}
    \label{fig:impact_scatter}
\end{figure}

All synthetic videos are rendered at $1920 \times 1080$ resolution with a fixed 20 FPS frame rate. Clip durations range from 5.8 to 32.2 seconds (mean 17.7 seconds, standard deviation 3.9 seconds). The ground-truth accident time occurs at a median of 6.9 seconds into the clip (interquartile range 5.2 to 9.8 seconds), which places most collisions in the first half of each recording (Figure~\ref{fig:time_dist}). Impact coordinates are normalized to $[0,1]^{2}$ and cluster toward the frame center: both $c_x$ and $c_y$ have means near 0.50 and standard deviations of 0.13 and 0.18, respectively (Figure~\ref{fig:impact_scatter}).

\subsection{Implementation}

We run inference on a single NVIDIA T4 GPU provided by the Kaggle platform. Collision classification uses the ViT-B/32 variant of CLIP~\citep{radford2021learning}. All video frames are resized to $320 \times 180$ for frame-differencing and optical flow. We do not train or fine-tune any model weights on either split of the dataset. Processing all 2,027 test videos takes approximately 2 hours.

Table~\ref{tab:hyperparams} lists the hyperparameters used across the pipeline. We chose these values by visual inspection on a handful of synthetic videos and kept them fixed for all test predictions.

\begin{table}[t]
\centering
\small
\caption{Pipeline hyperparameters, held constant for all test videos.}
\label{tab:hyperparams}
\begin{tabular}{@{}lll@{}}
\toprule
\textbf{Component} & \textbf{Parameter} & \textbf{Value} \\
\midrule
\multirow{2}{*}{Temporal}
  & Smoothing window $w$ & 5 \\
  & Z-score threshold $\tau$ & 1.5 \\
\midrule
\multirow{6}{*}{Spatial}
  & Start frame & centered on $t^{*}$ \\
  & Context window $T$ & 30 frames \\
  & Pyramid scale & 0.5 \\
  & Pyramid levels & 3 \\
  & Window size & 15 \\
  & Flow percentile threshold & 90th \\
\midrule
\multirow{2}{*}{Classification}
  & CLIP backbone & ViT-B/32 \\
  & Peak-region frames & 8 \\
\bottomrule
\end{tabular}
\end{table}

\subsection{Scoring}

Submissions are scored with the harmonic mean of three quantities. The temporal component $\mathcal{T}$ measures the proximity of the predicted accident time to the ground-truth via a Gaussian kernel with $\sigma_t = 2.0$ seconds: $\mathcal{T} = \exp\!\big({-\tfrac{1}{2}}\big(\tfrac{t_{\text{pred}} - t_{\text{gt}}}{\sigma_t}\big)^{2}\big)$. The spatial component $\mathcal{S}$ applies the same functional form with $\sigma_s = 0.1$ to the Euclidean distance between predicted and true impact coordinates in normalized $[0,1]$ space. The classification component $\mathcal{C}$ is top-1 accuracy: 1 if the predicted type matches the ground truth, 0 otherwise. The final score is $\mathcal{H} = 3 / (1/\mathcal{T} + 1/\mathcal{S} + 1/\mathcal{C})$, so a zero in any component forces the composite score to zero.

\subsection{Results}

Our pipeline achieves a public leaderboard score of \textbf{0.2523} on the real CCTV test set. This score is computed on approximately 25\% of the test data; the final ranking uses the remaining 75\%, so standings may shift.

On a 10-video calibration subset drawn from the synthetic split, the per-component mean scores are $\overline{\mathcal{T}} = 0.438$, $\overline{\mathcal{S}} = 0.168$, and $\overline{\mathcal{C}} = 0.0$. The zero classification score on this calibration subset is expected: all 10 calibration videos belong to the head-on category, but CLIP predicts t-bone for each one. The best individual temporal score (0.94) and the best individual spatial score (0.96) show that when the pipeline locks onto the correct event, both the time and location estimates can be accurate. The composite score drops to zero because the harmonic mean penalizes any single component failure.

\begin{figure}[t]
    \centering
    \includegraphics[width=\linewidth]{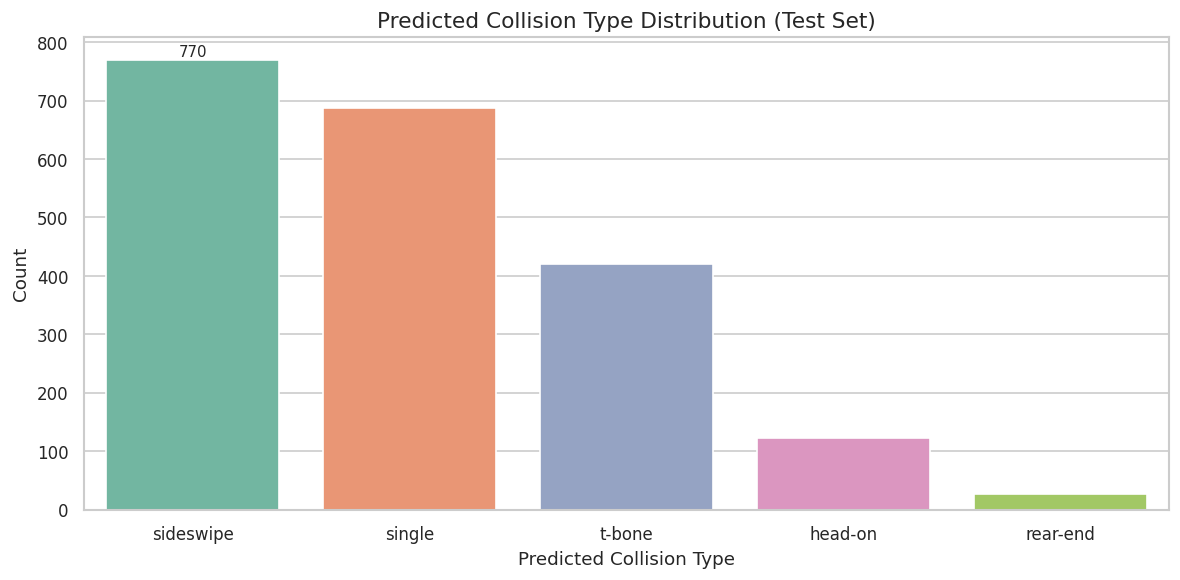}
    \caption{Predicted collision type distribution on the 2,027 real test videos. Sideswipe and single dominate the predictions, while rear-end is rarely selected, inverting the synthetic training distribution.}
    \label{fig:pred_dist}
\end{figure}

Figure~\ref{fig:pred_dist} shows the distribution of predicted collision types across the full test set. Sideswipe is the most frequently predicted category (770 of 2,027 videos), followed by single (687), t-bone (425), head-on (122), and rear-end (23). This distribution differs substantially from the synthetic training split (Figure~\ref{fig:type_freq}), where rear-end dominates. The discrepancy indicates that CLIP's visual similarity scores are sensitive to the viewing angle and scene geometry differences between CARLA renders and real CCTV footage, rather than reflecting the underlying collision dynamics.

\paragraph{Error analysis.}
Three error patterns dominate the pipeline. (1) Temporal localization breaks when the camera captures large-scale background motion, such as swaying vegetation or cloud shadows, that produces frame-difference spikes comparable in magnitude to actual collisions. (2) The spatial centroid drifts when multiple vehicles move simultaneously across the frame, because the weighted average spreads over all active regions instead of concentrating on one cluster. The ground-truth impact distribution (Figure~\ref{fig:impact_scatter}) clusters near the frame center, which partially explains why even noisy spatial predictions stay within a reasonable distance of the correct location. (3) CLIP misclassifies when CCTV viewpoints are overhead or oblique, since its training data consists of internet photographs taken at roughly eye level. The predicted type distribution (Figure~\ref{fig:pred_dist}) suggests that CLIP responds to general scene geometry cues rather than collision-specific visual features, which accounts for the systematic shift away from rear-end toward sideswipe and single categories.

\section{Conclusion}
\label{sec:conclusion}

We described a modular zero-shot pipeline for detecting, localizing, and classifying traffic accidents in CCTV footage, built for the ACCIDENT @ CVPR 2026 competition. The temporal module isolates the collision frame through z-score peak detection on frame differences. The spatial module accumulates Farneback optical flow, applies 90th-percentile thresholding, and returns the weighted centroid as the predicted impact point. The classification module matches CLIP image embeddings of peak-region frames against multi-prompt text embeddings to select a collision type. The pipeline achieves a public leaderboard score of 0.2523 without any domain-specific fine-tuning.

Because the three modules share no parameters, each can be upgraded on its own. Two directions seem most promising. First, replacing Farneback optical flow with a learned estimator like RAFT~\citep{teed2020raft} could improve displacement accuracy on low-resolution input. Second, fine-tuning the CLIP visual encoder on the synthetic split may narrow the domain gap between internet imagery and surveillance stills, which currently limits classification performance and represents the primary bottleneck identified in our calibration analysis.

\bibliographystyle{unsrtnat}
\bibliography{references}

\end{document}